# The Cyborg Astrobiologist: Testing a Novelty-Detection Algorithm on Two Mobile Exploration Systems at Rivas Vaciamadrid in Spain and at the Mars Desert Research Station in Utah

P.C. McGuire[1,2,3,10], C. Gross[1], L. Wendt[1], A. Bonnici[4], V. Souza-Egipsy[2,5], J. Ormö[2], E. Díaz-Martínez[2,7], B.H. Foing[6], R. Bose[3], S. Walter[1], M. Oesker[8,9], J. Ontrup[8], R. Haschke[8], H. Ritter[8]

[1]Freie Univ. Berlin, Germany
[2]Centro de Astrobiología (CSIC/INTA), Torrejón de Ardoz, Spain
[3]McDonnell Center for the Space Sciences, Washington Univ., St. Louis, USA
[4]Dept. Systems and Control Engineering, Univ. Malta
[5]currently at: Univ. Málaga, Spain
[6]ESTEC, Noordwijk, Netherlands
[7]currently at: Geological Survey of Spain (IGME), Madrid, Spain
[8]Technische Fakultät, Univ. Bielefeld, Germany
[9]currently at: IfG - Institute for Scientific Instruments GmbH, Berlin, Germany
[10]currently at: Dept. of Geophysical Sciences, Univ. Chicago; Email:mcguire@geosci.uchicago.edu

**Abstract**

In previous work, a platform has been developed for testing computer-vision algorithms for robotic planetary exploration (McGuire *et al.* 2004b,2005). This platform consisted of a digital video camera connected to a wearable computer for real-time processing of images at geological and astrobiological field sites. The real-time processing included image segmentation and the generation of interest points based upon uncommonness in the segmentation maps. Also in previous work, this platform for testing computer-vision algorithms has been ported to a more ergonomic alternative platform, consisting of a phone-camera connected via GSM network to a remote-server computer (Bartolo *et al.* 2007). The wearable-computer platform has been tested at geological and astrobiological field sites in Spain (Rivas Vaciamadrid and Riba de Santiuste), and the phone-camera has been tested at a geological field site in Malta. In this work, we (i) apply a Hopfield neural-network algorithm for novelty detection based upon color, (ii) integrate a field-capable digital microscope on the wearable computer platform, (iii) test this novelty detection with the digital microscope at Rivas Vaciamadrid, (iv) develop a Bluetooth communication mode for the phone-camera platform, in order to allow access to a mobile processing computer at the field sites, and (v) test the novelty detection on the Bluetooth-enabled phone-camera connected to a netbook computer at the Mars Desert Research Station in Utah. This systems engineering and field testing have together allowed us to develop a real-time computer-vision system that is capable, for example, of identifying lichens as novel within a series of images acquired in semi-arid desert environments. We acquired sequences of images of geologic outcrops in Utah and Spain consisting of various rock types and colors to test this algorithm. The algorithm robustly recognized previously-observed units by their color, while requiring only a single image or a few images to learn colors as familiar, demonstrating its fast learning capability.

## Introduction and Background

Currently, two robotic rovers are exploring Mars, having arrived there in January 2004 (Squyres *et al.* 2004a-b; Alexander 2006; Matthies *et al.* 2007). Two rovers are planned to be launched in the next few years, in order to explore Mars further (Barnes *et al.* 2006; Griffiths *et al.* 2006; Volpe 2003). Fink *et al.* (2005), Dubowsky *et al.* (2005), Yim, Shirmohammadi and Benelli (2007) and others suggest mission concepts for planetary or astrobiological exploration with 'tier-scalable reconnaissance systems', microbots, and marine/mud-plain robots, respectively[1]. Volpe (2003), Fink *et al.* (2005), Castano *et al.* (2007), Halatci, Brooks and Iagnemma (2007, 2008), and others have explored concepts within the area of scientific autonomy for such robotic rovers or missions on Mars and other planets or moons. We report upon progress made by our team in the development of scientific autonomy for the computer-vision systems for the exploration of Mars by robots. This work can also be applied to the exploration of the Moon, and the deserts and oceans of the Earth, either by robots or as augmented-reality systems for human astronauts or aquanauts. One motivation of our work is to develop systems that are capable of recognizing (in real time) all the different areas of a series of images from a geological field site. Such recognition of the complete diversity can be applied in multiple endeavors, including: image downlink prioritization, target selection for *in situ* analysis with geochemical or biological analysis tools, and sample selection for sample return missions.

Previous work by our team includes: (a) McGuire *et al.* (2004b) on the concept of uncommon maps based upon image segmentation, on the development of a real-time wearable computer system (using the NEO graphical programming language (Ritter *et al.* 1992,2002)) to test these computer-vision techniques in the field, and on geological/astrobiological field tests of the system at Rivas Vaciamadrid near Madrid, Spain; (b) McGuire *et al.* (2005) on further tests of the wearable computer system at Riba de Santiuste in Guadalajara, Spain – this astrobiological field site was of a totally different character than the field site at Rivas Vaciamadrid (red-bed sandstones instead of gypsum-bearing outcrops), yet the computer vision system identified uncommon areas at both field sites rather well; and (c) Bartolo *et al.* (2007) on the development of a phone-camera system for ergonomic testing of the computer-vision algorithms.

Previous work to ours includes: (A) McGreevy (1992, 1994) on the concept of geologic metonymic contacts, on cognitive ethnographic studies of the practice of field geology, and on using head-mounted cameras and a virtual-reality visor system for the ethnographic studies of field geologists[2]; (B) Gulick *et al.* (2001, 2004) on the development of autonomous image analysis for exploring Mars; and (C) Itti, Koch and Niebur (1998), Rae, Fislage and Ritter (1999), McGuire *et al.* (2002), Ritter *et al.* (2002), Sebe *et al.* (2003), and others on the development or testing of saliency or interest maps for computer vision. Current, related work to ours includes that of Purser *et al.* (2009) and Ontrup *et al.* (2009) on using Web 2.0 to label and explore images from a deep-sea observatory, using textural features and self-organizing maps.

---

[1] In this context, 'tier-scalable' means approximately 'multi-level', with (a) reconnaissance systems in orbit, (b) rovers and/or microbots on the ground, and (c) floating or flying robots in the atmosphere.
[2] By 'metonymic', McGreevy (1992,1994) refers to contrasting geological units, as opposed to 'metaphoric', which he would refer to units that are similar instead of contrasting. By 'ethnographic', McGreevy (1992,1994) studies field geologists somewhat like a 'tribe', in a sociological manner.

Herein, we extend our previous work by: (i) developing and testing a second computer-vision algorithm for robotic exploration by applying a Hopfield neural network algorithm for novelty detection (Bogacz, Brown and Giraud-Carrier, 1999, 2001), (ii) integrating a digital microscope into the wearable computer system, (iii) testing the novelty-detection algorithm with the digital microscope on the gypsum outcrops at Rivas Vaciamadrid, (iv) developing a Bluetooth communications mode for the phone-camera system, and (v) testing the novelty-detection algorithm with the Bluetooth-enabled phone-camera connected to an ordinary netbook computer (Wendt *et al.* 2009, Gross *et al.* 2009, Foing *et al.* 2009a-c) at the Mars Desert Research Station (MDRS) in Utah, which is sponsored and maintained by the Mars Society.

# Techniques

### NEO graphical programming language

NEO/NST is a graphical programming language developed at Bielefeld University (Ritter *et al.* 1992,2002) whose utilization (McGuire *et al.* 2004a) for the Cyborg Astrobiology project was based primarily upon its graphical representation and readability, promoting ease of use, debugging, and reuse. Previous computer-vision projects in Bielefeld, which use NEO, include the GRAVIS robot[3] (McGuire *et al.* 2002; Ritter *et al.* 2003). The NEO language includes icons for each algorithmic subroutine, wires that send data to and from different icons, encapsulation to create meta-icons from sub-icons and wires, capability to code icons either in compiled or interpreted C/C++ code, and capabilities for iterative real-time debugging of graphical sub-circuits. NEO functions on operating systems which include Linux and Microsoft Windows. For the field-work for this project, the Microsoft Windows version of NEO was used exclusively. However, much of the developmental computer-vision programming was done at the programmer's desk using the Linux version of NEO.

### Image segmentation

In 2002, three of the authors (Ormö, Díaz-Martínez and McGuire) decided that image segmentation is a low-level technique that was needed for the Cyborg Astrobiologist project, in order to distinguish between different geological units, and to delineate geological contacts. McGreevy (1992, 1994) naturally also considers geologic metonymic contacts to be an essential cognitive concept for the practice of field geology. Hence, a gray-scale image segmentation meta-icon was programmed from a set of preexisting compiled NEO icons together with a couple of newly-created compiled NEO icons.

This image segmentation meta-icon is based upon a classic technique of using two-dimensional co-occurrence matrices (Haralick 1973) to search for pairs of pixels in the image with common gray-scale values to other pairs of pixels.[4] Six to eight peaks in the two-dimensional co-occurrence matrix are identified using a NEO icon, and all pixels in each of these peaks is labeled as belonging to one segment or cluster of the gray-scale image. Therefore, we are provided with 6-8 segments for each image; see McGuire *et al.* (2004b, 2005) for more details.

---

[3] The **G**estural **R**ecognition **A**ctive **VI**sion **S**ystem (GRAVIS) robot is a vision-based robot developed in Bielefeld, Germany, which has a hybrid architecture that combines statistical methods, neural networks, and finite state machines into an integrated and flexible framework for controlling its behavior.

[4] Alternatively, we could have used one-dimensional histograms to identify single pixels in the image with common gray-scale values to other pixels. However, the two-dimensional co-occurrence matrices added value to the project since they also offer the abilities to identify geological contacts and to characterize texture. Nonetheless, in the current system, the capabilities for identifying texture and contacts are largely unutilized.

For the work described in this paper, which involves color novelty detection, we needed to identify regions of relatively homogeneous color (in RGB or HSI color space[5]), as opposed to relatively homogenous gray-scale values. So we implemented a full-color image segmentation algorithm in NEO, based upon three-band spectral-angle matching. This spectral angle matching thus provides full-color image segmentation, a capability which was lacking in previous work within this project. In our implementation of spectral angle matching, each pixel is represented as a three-vector[6] with its Hue (H), Saturation (S) and Intensity (I) values. This three-vector for a given pixel is compared to the three-vectors for other pixels with elementary trigonometry by computing the angle between these two three-vectors. If this angle is sufficiently small, the two pixels are assigned to be in the same color segment or cluster. Otherwise, they are assigned to different color segments or clusters. By evaluating the spectral angle between each unassigned pixel pair in the image, making assignments sequentially, each pixel is assigned to be in a color segment. The matching angle is set before the mission but can be modified if desired during the mission. This technique is commonly used in multispectral and hyperspectral image classification, so the novelty detection (which we describe later) which uses this spectral-angle matching technique for color image segmentation is, in principle, directly extensible to multispectral imaging.

### Interest points by uncommon mapping

Though this is not the focus of the current paper, this exploration algorithm of 'interest points by uncommon mapping' was tested further during both field missions described in this paper. In previous field missions, this was the only algorithm which was tested and evaluated (McGuire *et al.* 2004b, 2005). In summary, an uncommon map is made for each of the three bands of a color HSI image, with the image segments that had the least number of pixels being given a numerical value corresponding to high uncommonness, and the image segments with the highest number of pixels being given a numerical value corresponding to low uncommonness. Three uncommon maps are the result of this process. The three uncommon maps are summed together to produce an interest or saliency map. The interest map is then blurred with a smoothing operator in order to search for clusters of interesting pixels instead of solitary interesting pixels. The three highest peaks in the blurred interest map were identified and reported as 'interest points' to the human user.

### Novelty detection

By utilizing the three-color image segmentation maps for an image, we compute mean values for the hue, saturation and intensity (<H>, <S>, <I>) for each segment of each image that is acquired. This vector of three real numbers is then converted to a vector of three 6-bit binary numbers in order to feed it into an 18-neuron Hopfield neural network, as developed for novelty detection by Bogacz, Brown and Giraud-Carrier (1999, 2001) and coded into NEO for this project. If the interaction energy of the 18-vector with the Hopfield neural network is high, then the pattern is considered to be novel and is then stored in the neural network. Otherwise, the pattern has a low familiarity energy, it is considered to be familiar (not novel), and is not stored in the neural network. The novelty-detection threshold is fixed at $-N/4 = -4.5$, as prescribed by Bogacz, Brown and Giraud-Carrier (1999, 2001). See Fig. 1 for a summary and Fig. 2 for an offline example computed from data from the red-beds in Riba de Santiuste in Guadalajara, Spain. The familiar segments of the segmented image are colored black, and the novel segments of the segmented image retain the colors that were used in the segmented image. In so doing, a quick inspection of the novelty maps allows the user to see what the computer considers to be novel (at least based upon this simple color prescription without texture). See Fig. 3 for an example of the real-time novelty detection with the digital microscope during the Rivas Vaciamadrid mission in September 2005. As shown in Fig. 3, red-colored sporing bodies of lichens were identified to be novel on the 18[th] image of the mission, and similar red-colored sporing bodies of lichens were found to be familiar on the 19[th] image of the mission. This is 'fast learning'.

---

[5] The representation of an RGB image instead with HSI provides certain advantages over the RGB representation. For example, the intensity component is separated naturally from the 'color' components (hue and saturation), thus providing a representation where the chromatic components can be clustered separately from the intensity component. This can be useful if there are variations in the flux of incident light or if there are shadows.

[6] A three-vector is defined as a vector with three components.

An alternative, complementary direction for the development of the Cyborg Astrobiologist project (in addition to the current project of novelty detection) would be to direct the computer to identify units of similar color to a unit of a user-prescribed color. When extended to multispectral or hyperspectral imaging cameras, this would allow the search for minerals of interest.

# Hardware

## Wearable Computer

In prior articles of this journal, the technical specifications of the wearable Cyborg Astrobiologist computer system have been descibed (McGuire *et al.* 2004b, 2005). It consists of PC hardware with special properties to be both power-saving and worn on the explorer's body. These special properties include: an energy-efficient computer, a keyboard and trackball attached to the human's arm, and a head-mounted display. We also use a color video camera connected to the wearable computer by a Firewire communications cable, which is mounted on a tripod. The mounting on the tripod was necessary for reducing the vibrations in the images and for increasing the repeatability of pointing the camera at the same location in the geological outcrops.

For the field experiments at Rivas Vaciamadrid discussed in this paper, the wearable-computer system has been extended by having the option to replace the color video camera with a USB-attached digital microscope.

## Digital Microscope

The digital microscope for the Rivas Vaciamadrid mission was chosen as an off-the-shelf USB microscope (model: Digital Blue QX5), which is often used for children's science experiments, but with sufficient capacity for professional scientific studies (e.g. King and Petruny 2008). This microscope has three magnifications (10x, 60x, 200x) and worked fine with the wearable computer, allowing contextual images of ~1 mm diameter sporing bodies of lichens. It imaged well in natural reflective lighting in the field. A small tripod was added to the microscope to facilitate alignment and focusing in the field. As a result of the field experience at Rivas Vaciamadrid in September 2005, we recommend that future preparatory work with the digital microscope should involve the development of a robotic alignment-and-focusing stage for the microscope. This would save considerable time in the field, whereas the alignment and focusing were done manually during the Rivas mission.

**Phone Camera**

The phone camera for the mission to the Mars Desert Research Station (MDRS) in Utah was chosen as an off-the-shelf mobile phone camera (model: Nokia 3110 classic, with 1.3 megapixels), which is used quite commonly in everyday personal and professional communication. This phone camera has Bluetooth capabilities as well as standard mobile phone communication capabilities via the mobile-phone communication towers. The camera in the phone camera has $480 \times 640$ pixels, and the images are transmitted by Bluetooth in lossy JPEG format. This particular camera turned out commonly to have a slight (<5-10%) enhancement in the central ~50% of the image (relative to the edges of the image) in the red channel of the RGB images. This issue was not noticed for the green and blue channels, and our novelty-detection software handled the red-channel enhancement rather robustly.

**Netbook Computer**

For the mission to the MDRS, a small netbook computer has been chosen for the automated processing of the phone-cam images. The "ASUS Eee PC 901" computer is noted for its combination of a light weight, power-saving processor, solid-state disk drive and relatively low cost. It features an Intel Atom "Diamondville" CPU clocked at 1.6 GHz, Bluetooth and 802.11n Wi-Fi connection. Proper power management software and a 6-cell battery enable battery life of up to 7 hours. The computer is "paired" with the mobile phone cam by the Bluetooth stack in such a way that the images are transferred automatically to a certain folder which is observed by the automation software. As the "Class 1" Bluetooth connection allows transmission distances of up to 100 m, the computer can either be carried by the human who is operating the phone-cam, by another nearby colleague, or be stationary in, for instance, a nearby vehicle.

**Bluetooth and automation**

The communications and automation software between the phone-cam and the processing computer can handle images with both a standard mobile phone network (Borg, Camilleri and Farrugia 2003; Bartolo *et al.* 2007) and via Bluetooth communications (this work). The Bluetooth communication and automation software between the phone camera and the computer (netbook) was programmed in the MATLAB language and then compiled for use in the field. The MATLAB program uses: (a) Microsoft Windows system commands and MATLAB internal commands to control manipulation and conversion for the image files; (b) menu macro commands to send and receive images via Bluetooth; and (c) a 'spawn call' of the NEO program for the computer vision. The MATLAB program allows the user to choose to use either the novelty detection or the interest-point detection from uncommon mapping, and it also allows the user to set the novelty detection neural-network memory to 'zero'. Future work with the Bluetooth automation may involve using a Bluetooth Application Programming Interface (API) instead of the MATLAB menu macro commands, in order to speed up the communications.

# Field Sites

## Rivas Vaciamadrid in Spain

On the 6th of September, 2005, two of the authors (McGuire and Souza-Egipsy) tested the wearable computer system for the fifth time at the Rivas Vaciamadrid geological site, which is a set of gypsum-bearing southward-facing stratified cliffs near the "El Campillo" lake of Madrid's Southeast Regional Park (Castilla Cañamero[7]; Calvo, Alonso and Garcia del Cura 1989; IGME 1975), outside the suburb of Rivas Vaciamadrid. This work involved a digital microscope and a novelty detection algorithm. Prior work at Rivas Vaciamadrid by our team was described in McGuire *et al.* 2004b:

> "The rocks at the outcrop are of Miocene age (15-23.5 Myrs before present), and consist of gypsum and clay minerals, also with other minor minerals present (anhydrite, calcite, dolomite, etc.). These rocks form the whole cliff face which we studied, and they were all deposited in and around lakes with high evaporation (evaporitic lakes) during the Miocene. They belong to the so-called Lower Unit of the Miocene. Above the cliff face is a younger unit (which we did not study) with sepiolite, chert, marls, etc., forming the white relief seen above from the distance, which is part of the so-called Middle Unit or Intermediate Unit of the Miocene. The color of the gypsum is normally grayish, and in-hand specimens of the gypsum contain large and clearly visible crystals.
>
> We chose this locality for several reasons, including:
> • It is a so-called 'outcrop', meaning that it is exposed bedrock (without vegetation or soil cover);
> • It has distinct layering and significant textures, which will be useful for this study and for future studies;
> • It has some degree of color differences across the outcrop;
> • It has tectonic offsets, which will be useful for future high-level analyses of searching for textures that are discontinuously offset from other parts of the same texture;
> • It is relatively close to our workplace;
> • It is not a part of a noisy and possibly dangerous construction site or highway site.
>
> The upper areas of the chosen portion of the outcrop at Rivas Vaciamadrid were mostly of a tan color and a blocky texture. There was some faulting in the more layered middle areas of the outcrop, accompanied by some slight red coloring. The lower areas of the outcrop were dominated by white and tan layering. Due to some differential erosion between the layers of the rocks, there was some shadowing caused by the relief between the layers. However, this shadowing was perhaps at its minimum for direct lighting conditions, since both expeditions were taken at mid-day and since the outcrop runs from East to West. Therefore, by performing our field-study at Rivas during mid-day, we avoided possible longer shadows from the sun that might occur at dawn or dusk."

## Mars Desert Research Station (MDRS) in Utah

For the last two weeks of February 2009, two of the authors (Gross and Wendt) tested the Bluetooth-enabled phone-camera system with the novelty detection algorithm at the MDRS field site in Utah (see Figs. 4-7). The MDRS is located in the San Rafael Swell of Utah, 11 Kilometers from Hanksville in a semi-arid desert. Surrounding the MDRS-Habitat, the Morrison Formation represents a sequence of Late Jurassic sedimentary rocks, also found in large areas throughout the western United States. The deposits are known to be a famous source of dinosaur fossils in North America. Furthermore, the Morrison Formation is a major source for uranium mining in the US. The formation is named after Morrison, Colorado, where Arthur Lakes discovered the first fossils in 1877. The sequence is composed of red, greenish, grey or light gray mudstone, limestone, siltstone and sandstone, partially inter-bedded by white sandstone layers. Radiometric dating of the Morisson Formation show ages of 146.8 +/- 1 Ma (top) to 156.3 +/- 2 Ma (base) (Bilbey 1998).

---

[7] Castilla Cañamero, G. (2001), "Informe sobre las prácticas profesionales realizadas en el Centro de Educación Ambiental: El Campillo", Internal report to the Consejería de Medio Ambiente de la Comunidad de Madrid.

The Morrison Formation is divided into four subdivisions. From oldest to most-recent they are: Windy Hill Member (shallow marine and tidal flat deposits), Tidewell Member (lake and mudflat deposits), Salt Wash Member (terrestrial deposition, semi arid alluvial plain and seasonal mudflats) and the Brushy Basin Member (river and saline/alkaline lake deposits). The Cyborg system was tested on two of the Members, the Salt Wash and the Brushy Basin Members, once deposited in swampy lowlands, lakes, river channels and floodplains. The Brushy Basin Member is much finer-grained than the Salt Wash Member and is dominated by mudstone rich in volcanic ash. The most significant difference of the Brushy Basin Member is the reddish color, whereas the Salt Wash Member has a light, grayish color. The deposits were accumulated by rivers flowing from the west into a basin that contained a giant, saline, alkaline lake called Lake T'oo'dichi' and extensive wetlands that were located just west of the Uncompahgre Plateau. The large alkaline, saline lake, Lake T'oo'dichi', occupied a large area of the eastern part of the Colorado Plateau during the deposition. The lake extended from near Albuquerque, New Mexico, to near the site of Grand Junction, Colorado, and occupied an area that encompassed the San Juan and ancestral Paradox basins, making it the largest ancient alkaline, saline lake known (Turner *et al.* 1991). The lake was shallow and frequently evaporated. Intermittent streams carried detritus from source areas to the west and southwest far out into the lake basin. Prevailing westerly winds carried volcanic ash to the lake basin from an arc region to the west and southwest (Kjemperud *et al.* 2008). Today, the rough and eroded surface of the Brushy Basin- and the Salt Wash Member in the area around the MDRS show a variety of different morphologic features, making it ideal for testing the Cyborg Astrobiologist System.

A brief summary of the characteristics of the Bushy Basin and Salt Wash Members of the Morrison Formation is given here:

- color variations within one unit due to changing depositional environment and/or volcanic input
- diverse textures of the sediment surface (popcorn-texture, due to bentonite)
- laminated sediment outcrops
- eroded boulder fields in the Salt Wash Member
- cross bedding in diverse outcrops
- interbedded gypsum nodules
- lichens and endolithic organisms
- scarce vegetation.

## Results

The main results of this development and testing are three-fold. First, we have two rather robust systems (the wearable computer system with a digital microscope and the Bluetooth-enabled netbook/phone-cam system) that have different complementary hardware capabilities. The wearable computer system can be used either when advanced imaging/viewing hardware is required or when the most rapid communication is required. The netbook/phone-cam system can be utilized when higher mobility or ease-of-use is required. Second, our novelty-detection algorithm and our uncommon-mapping algorithm are also both rather robust, and ready to be used in other exploration missions here on the Earth, as well as being ready to be optimized and refined further for exploration missions on the Moon and Mars. Such further optimizations and refinements could include: optimization for speed both of computation and of communication, the incorporation of texture (for example, Freixenet *et al.* 2004) into the novelty-detection and uncommon-mapping, the extension to more than three bands of color information (i.e., multispectral image sensors), and improvements to the clustering used in the image segmentation module, etc. Third, the NEO graphical-programming language is a mature and easy-to-use language for programming robust computer-vision systems. This observation is based upon over eight years of experience of using NEO on this project (McGuire *et al.* 2004a-b,2005; Bartolo *et al.* 2007, and the current work) and on the predecessor GRAVIS project (McGuire *et al.* 2002; Ritter *et al.* 2002).

### Results at Rivas Vaciamadrid

We show one result for Rivas Vaciamadrid in Fig. 3. The deployment of the digital microscope is shown on the far left.

The four rows of the $4 \times 4$ sub-array of images represent selected images from the mission to Rivas Vaciamadrid, with the earliest images at the top and the latest images at the bottom.

The four columns of the $4 \times 4$ sub-array of images (from left to right) are:

1. the original acquired images from the panoramic camera and the digital microscope;
2. the full-color image segmentation (using spectral angle matching);
3. uncommon maps from the full-color image segmentation;
4. novelty maps from the sequence of full-color segmented images.

Of particular note from this series of images are the $2^{nd}$-$4^{th}$ rows of this sub-array of images. In the $2^{nd}$ row, corresponding to the $18^{th}$ image in the image sequence, we have imaged the red sporing-bodies of a lichen for the first time during this mission. Each of the sporing-bodies is approximately 1 mm in diameter. The uncommon map for this image finds the sporing-bodies to be relatively uncommon colored regions of the image. Impressively, the novelty map for this image finds the red sporing-bodies to be the nearly-unique areas of this image that are novel. The other area that was determined to be novel was a small area of brownish rock in the upper right hand corner of the image, beneath the white-colored parent-body of the lichen. More impressively, the novelty map for the image in the $3^{rd}$ row, in the $19^{th}$ image in the image sequence, does not find these red sporing-bodies to be novel. The red sporing-bodies have been observed previously, so the colors of the sporing-bodies are now familiar to the Hopfield neural network. The neural network has learned these colors after only one image! In the $4^{th}$ row ($23^{rd}$ image of the image sequence), the Hopfield neural network detects some black and orange lichens as being novel as well.

### Results at the Mars Desert Research Station

We show some of the results for the mission to the MDRS in Figs. 8-12. The first images were taken at typical outcrops of the sediments with so-called popcorn structures (see Fig. 8). These structures are caused by the presence of bentonite in the sediments. Small channels exposed a cross-section through the stratigraphic sequence, so we imaged the ground through these diverse small channels and further up the geologic sequence. The system robustly detected color changes from gray to red and white, with the novelty map lighting up when new colors were observed. In addition to this, the system recognized different structures as being novel, such as boulders of different colors and shadows. Generally, a terrain of uniform color was completely familiar to the novelty-detection system after 3 to 6 pictures (see Fig. 9). Objects whose colors were not familiar to the system were detected as being novel without exception (Figs. 9-11). By far the best results of the Cyborg Astrobiologist were obtained from lichens and cross-cutting layerings of gypsum or ash (Fig. 10-11). Lichens were detected very accurately as being novel, whether they were in the shadow or in bright sunlight, even after some prior observations of lichens of similar colors. During the tests, a library of pictures was developed and categorized for further investigations and instrument tests, for example in processing-quality and recognition experiments. The library consists of 259 images of the desert landscape, most of them without vegetation and some with soils at different humidity levels. A subset of thumbnails of images from this library is shown in Fig. 8.

The novelty detection system robustly separated images of rock or biological units it had already observed before from images of surface units which it had not yet observed. Sometimes, several images are required in order to learn that a certain color has been observed before (see Fig. 11). Nonetheless, this shows that color information can be used successfully to recognize familiar surface units, even with rather simple camera systems like a mobile-phone camera. For remotely operating exploration systems, this ability might be a great advantage due to its simplicity and robustness.

The hardware and software performed very well and was very stable. Some problems emerged when using the system with big gloves in the EVA-suit and when reading out the phone-cam display in the bright light. Some problems emerged from cast shadows in rough terrain or at low standing sun. The system detected those sharp contrast transitions as novel.

The uncommon-mapping system for the generation of interest points (McGuire *et al.* 2004b, 2005) was also tested again at a few locations. We targeted broad scenes (see Fig. 12) in order to test the landscape capability of the system with the goal of identifying, for example, obstacles or layer boundaries which could stand in the way of a rover.

## Summary and Future Work

We have developed and tested a novelty-detection algorithm for robotic exploration of geological and astrobiological field sites in Spain and Utah, both with a field-capable digital microscope connected to a wearable computer and a hand-held phone-camera connected to a netbook computer via Bluetooth.

The novelty-detection algorithm can detect, for example, small features such as lichens as novel aspects of the image sequence in semi-arid desert environments. The novelty-detection algorithm is currently based upon a Hopfield neural network that stores average HSI color values for each segment of an image segmented with full-color segmentation. The novelty-detection algorithm, as implemented here, can learn familiar color features sometimes in single instances, though several instances are sometimes required. The algorithm never requires many instances to learn familiar color features.

In many machine-learning applications, learning a piece of information often requires the presentation of multiple, or even many, instances of similar pieces of information, though single-instance learning is possible in certain situations with particular algorithms (Aha, Kibler, and Albert 1991; Maron and Lozano-Perez 1998; Zhang and Goldman 2002; Chen and Wang 2004).

Humans are sometimes capable of learning a new piece of information with only a single instance (Read 1983), though multiple instances are often required (Gentner and Namy 2006). Our implementation of the novelty-detection algorithm (Bogacz, Brown and Giraud-Carrier 1999, 2001) is therefore competitive with both humans and other machine-learning systems, at least in terms of training requirements, for this application of the real-time detection of unfamiliar colors in images from a semi-arid desert.

In our next efforts, we intend to (a) extend this processing to include robust segmentation, based upon both color and texture (for example, Freixenet *et al.* 2004), as well as novelty detection based upon both color and textural features; (b) optimize the current system for speed of computation and communication; and (c) extend our system to more sophisticated cameras (i.e., multispectral cameras).

In the near future, we will test the Bluetooth-enabled phone-camera and novelty-detection system at additional field sites with different types of geological and astrobiological images than we have studied thus far.

## Acknowledgements:

We would like to acknowledge the support of other research projects which helped in the development of the Cyborg Astrobiologist. Integration of the camera-phone with an automated mail watcher was carried out under the `Innovative Early Stage Design Product Prototyping' (InPro) project, supported by the University of Malta under research grant IED 73-529-2005. Many of the extensions to the GRAVIS interest-map software, programmed in the NEO language, were made as part of the Cyborg Astrobiologist project from 2002-2005 at the Centro de Astrobiologia in Madrid, Spain, with support from INTA and CSIC, and from the Spanish Ramon y Cajal program.

We thank the Mars Society for access to their unique facility in Utah, ESA/ESTEC for facilitating the mission at the Mars Desert Research Station in February 2009, and Gerhard Neukum for his support of the Utah mission in February 2009.

PCM acknowledges support from a research fellowship from the Alexander von Humboldt Foundation, as well as from a Robert M. Walker fellowship in Experimental Space Sciences from the McDonnell Center for the Space Sciences at Washington University in St. Louis.

PCM is grateful for conversations with: (a) Peter Halverson, which were part of the motivation for developing the Astrobiology Phone-cam; (b) Javier Gómez-Elvíra and Jonathan Lunine, which were part of the motivation for developing the Cyborg Astrobiologist project; and (c) Babette Dellen, concerning the novelty-detection neural-network algorithm.

CG and LW acknowledge support from the Helmholtz Association through the research alliance "Planetary Evolution and Life".

Novelty Detection w the
FamE (FamiliarityEnergy) Hopfield Neural
Net (R. Bogacz, U. Bristol) (using NEO)

Compute:

    averages: $<H>, <S>, <I>$ for each segment

    $x = (<H>, <S>, <I>)^T$    ;;vector=3 numbers with
            ;;6 bits/number = a 18 bit binary vector

If (Energy_Hopfield( x ) < -N/4 then          ;;N=18

    Novel = FALSE;

Else {

    Novel = TRUE;

    Store Pattern  x   in Hopfield Net;  }

Repeat for all segments in incoming image;

**Figure 1:** High-level pseudo-program for the current novelty-detection module of the Cyborg Astrobiologist. The number of neurons in the Hopfield neural network is N=18, corresponding to a threshold of -4.5 for familiarity energy.

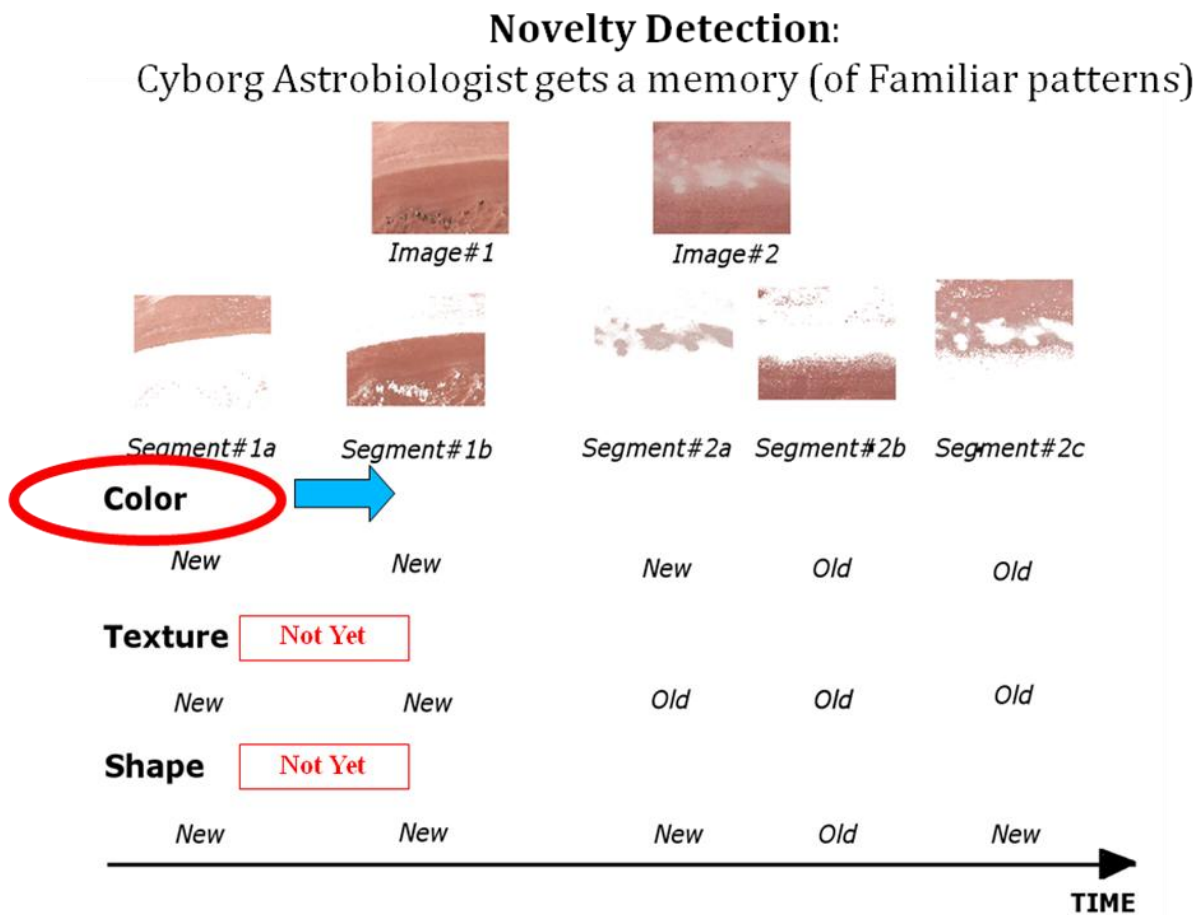

**Figure 2:** An example of color-based novelty detection, using two images from an image sequence from a previous mission to Riba de Santiuste, processed in an offline manner with color-based image segmentation and color-based novelty detection. We have not yet implemented texture- and shape-based novelty detection. These images from Riba de Santiuste are approximately 0.30 m wide.

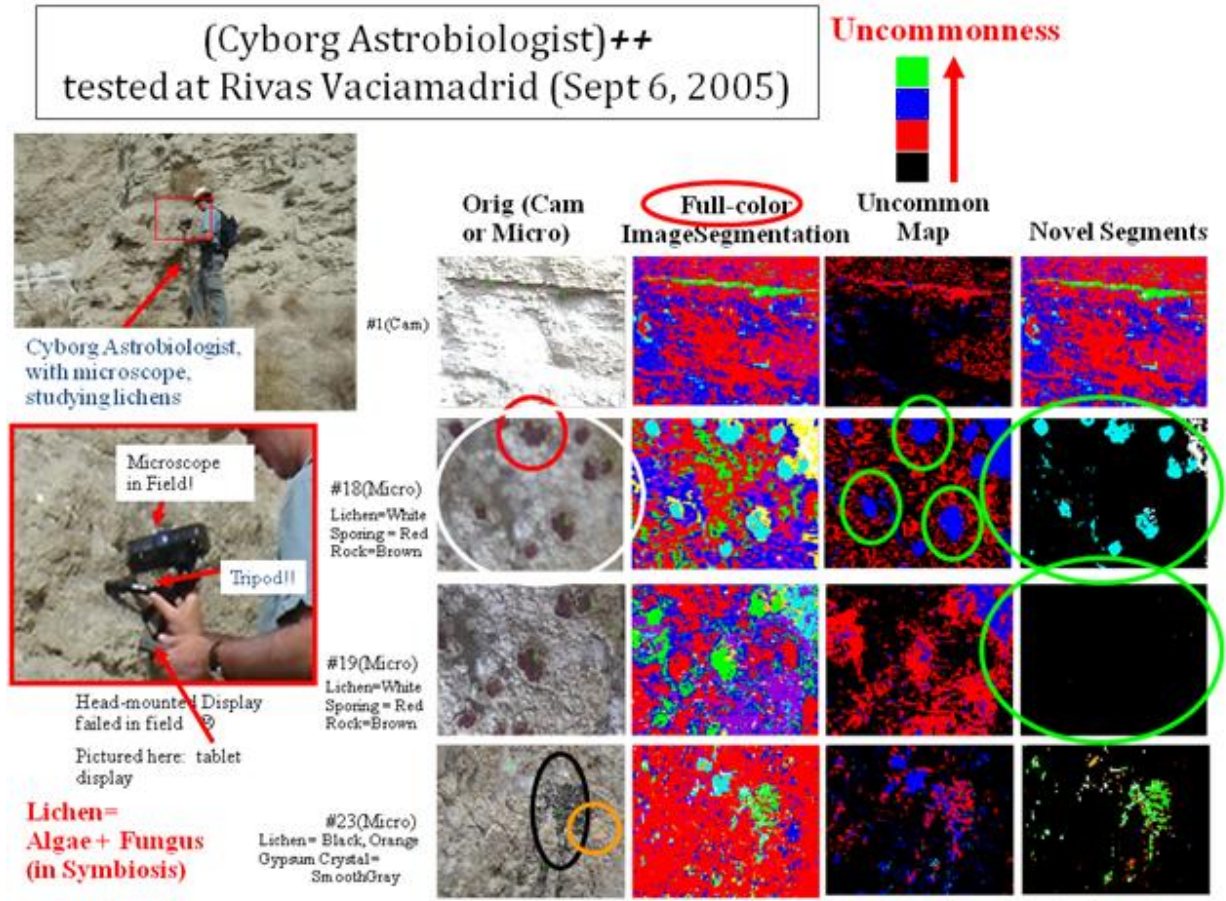

**Figure 3:** Demonstration of real-time color-based novelty detection with a field-capable digital microscope at Rivas Vaciamadrid. Image#1 was acquired by the panoramic camera prior to switching the input to be from the digital microscope, and this image corresponds to a cliff face exposure of about 10 m in width. The bottom three images acquired with the digital microscope (images #18, #19, and #23) are about 0.01 m wide. In image #18, the sporing-bodies of lichens are identified in image#18 as being novel, whereas in image #19, similar red sporing-bodies are regarded as familiar. Black and orange lichens are also regarded as novel in image #23. See the text for discussion.

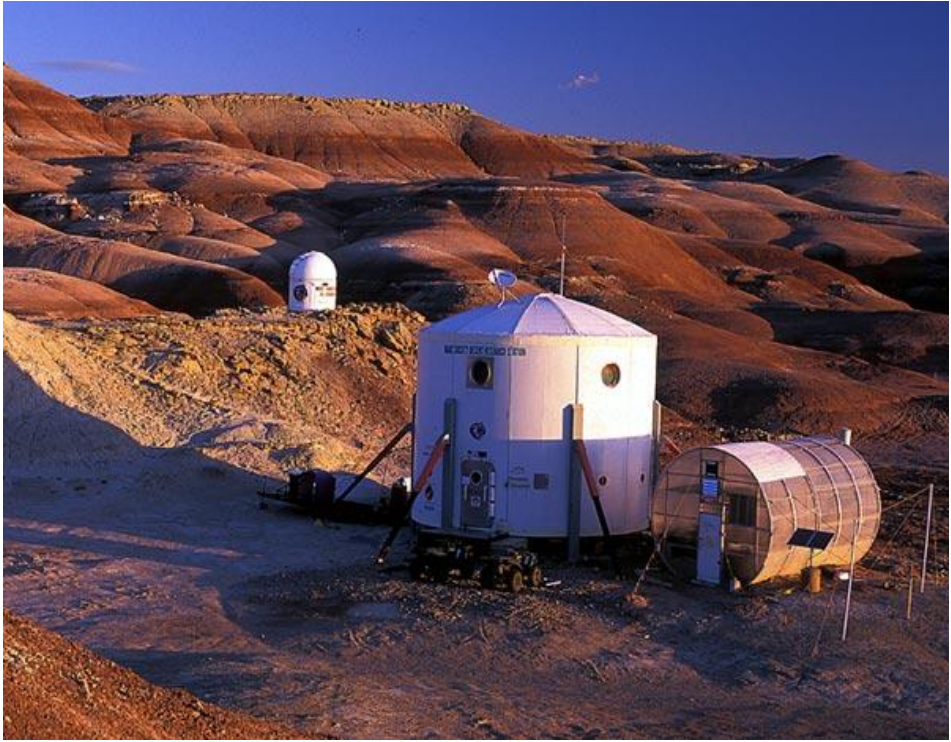

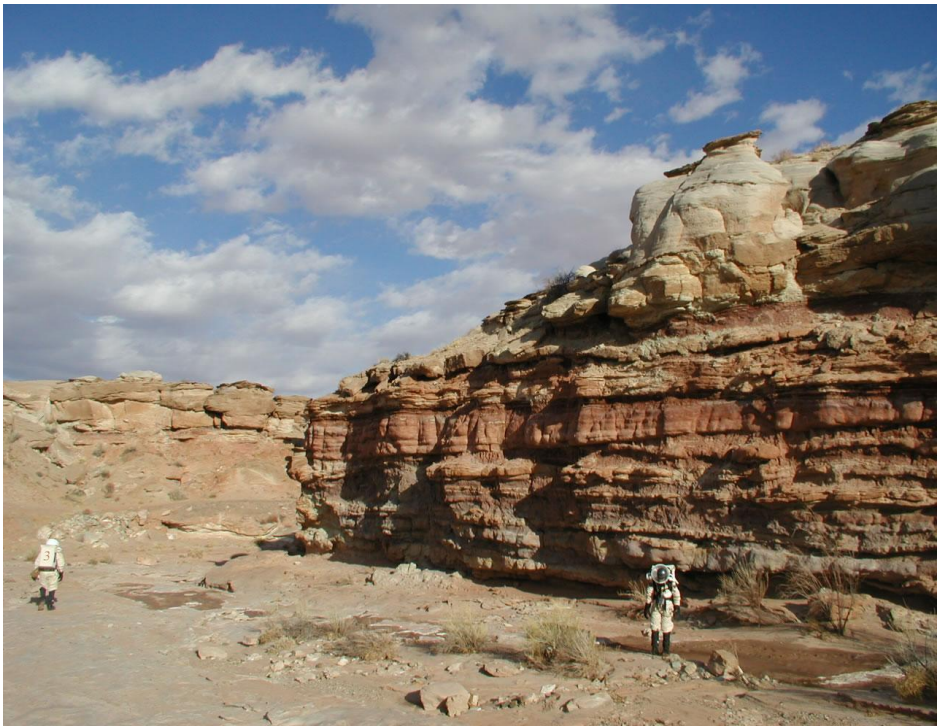

**Figure 4: (top)** The Mars Desert Research Station in Utah. **(bottom)** Two 'astronauts' exploring a stream near the MDRS.

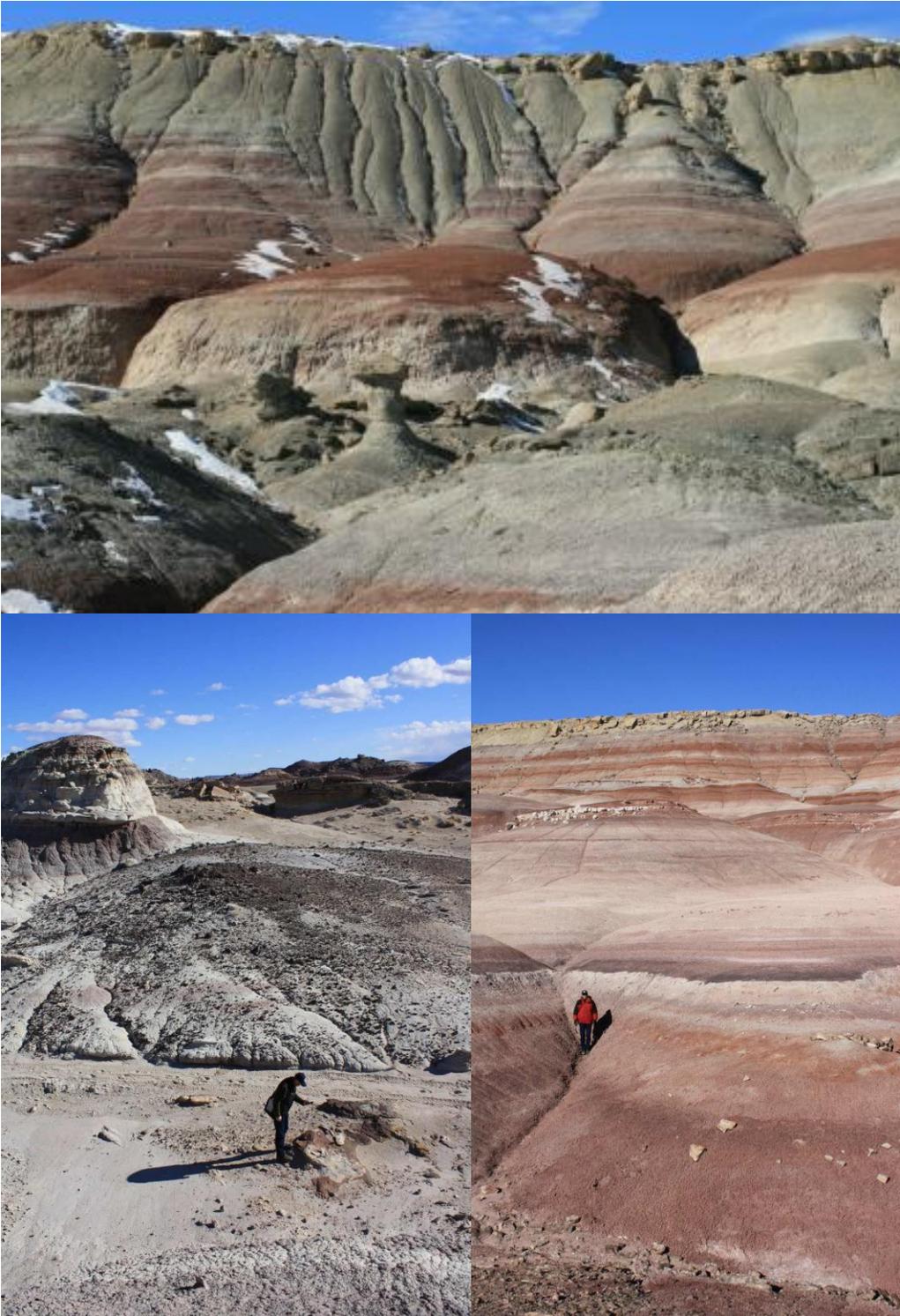

**Figure 5:** (Top) Panoramic view of the Morrison cliffs near the MDRS. (Bottom) Two images of the Cyborg Astrobiologist exploring the Morrison cliffs equipped with a Bluetooth-enabled phone-cam and a netbook. Note the impressive mobility (relative to many robotic astrobiologists) in the gully on the right image.

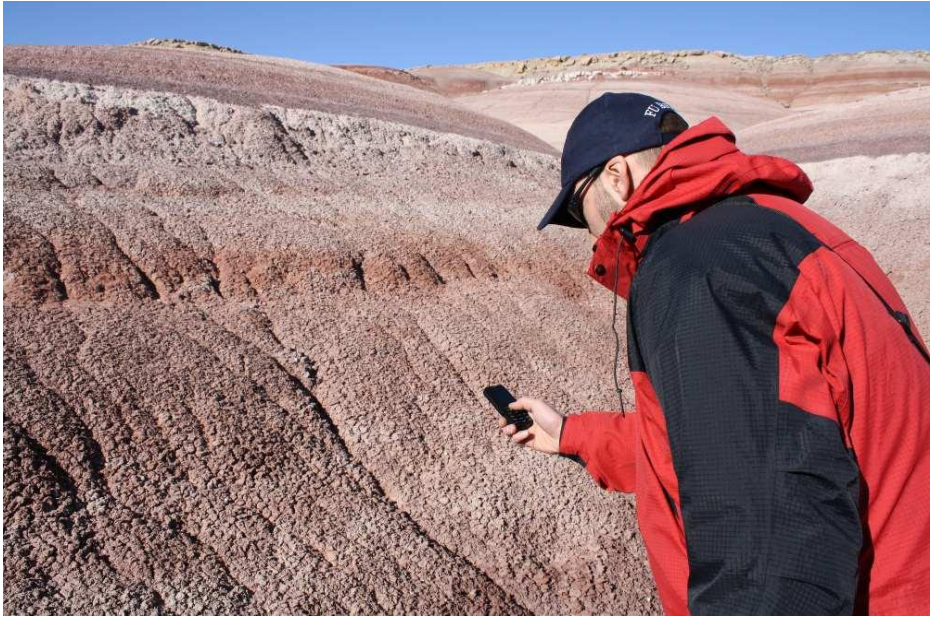

a)

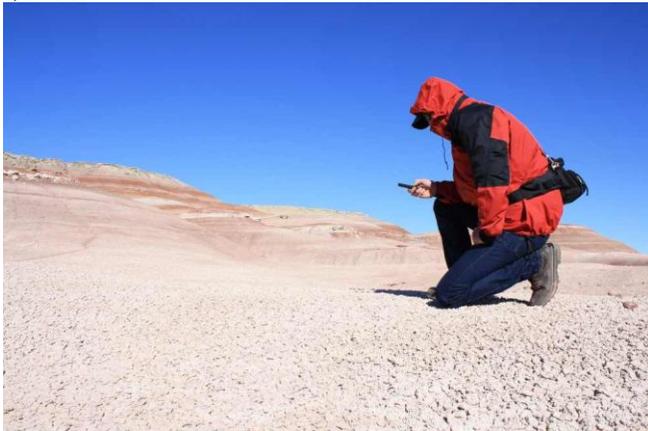

b)

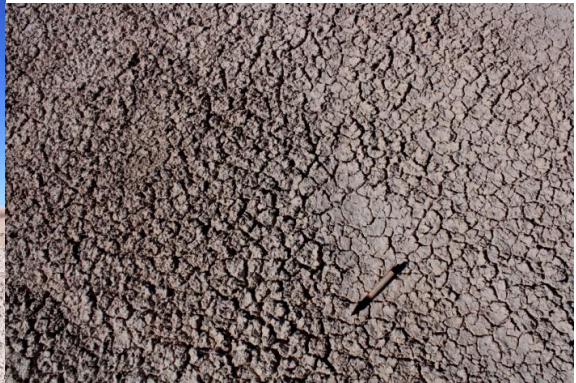

c)

**Figure 6:** Two close-up views (Subfigs. a-b) of the Cyborg Astrobiologist in action at the MDRS, together with an image (Subfig. c) acquired by the Cyborg Astrobiologist with the Astrobiology Phone-cam at the location in Subfig. b. In other tests at the MDRS, the Cyborg Astrobiologist when suited as an astronaut was able to successfully, but less robustly, operate the Bluetooth-enabled phone-camera (phone-camera operation was limited by the big finger problem of the gloved astronaut).

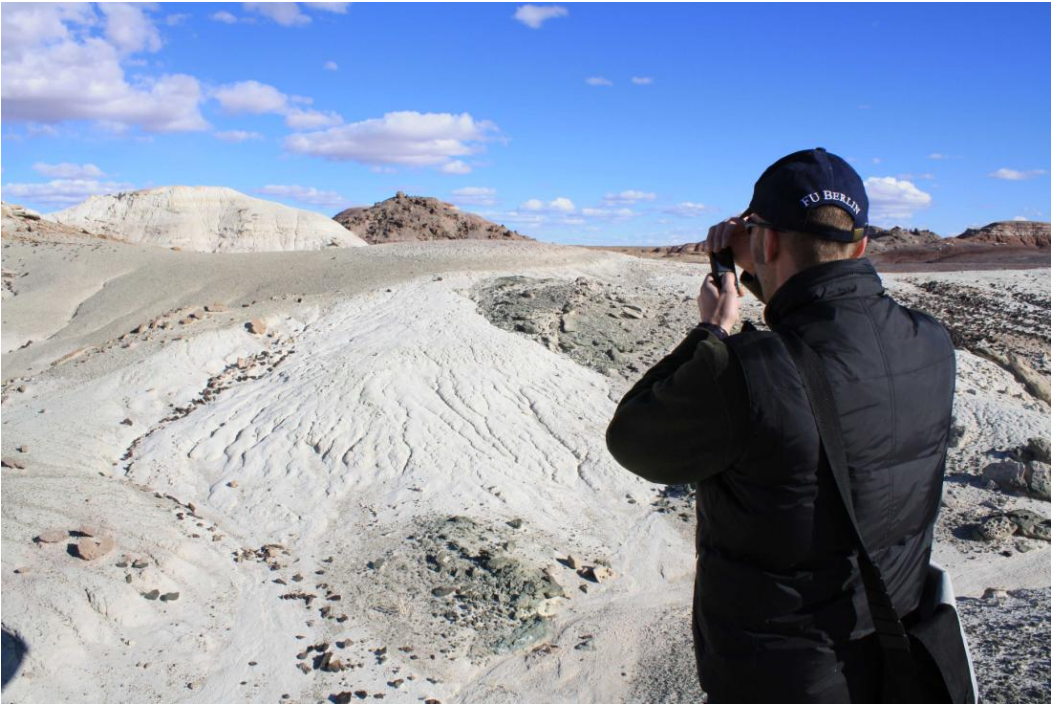

a)

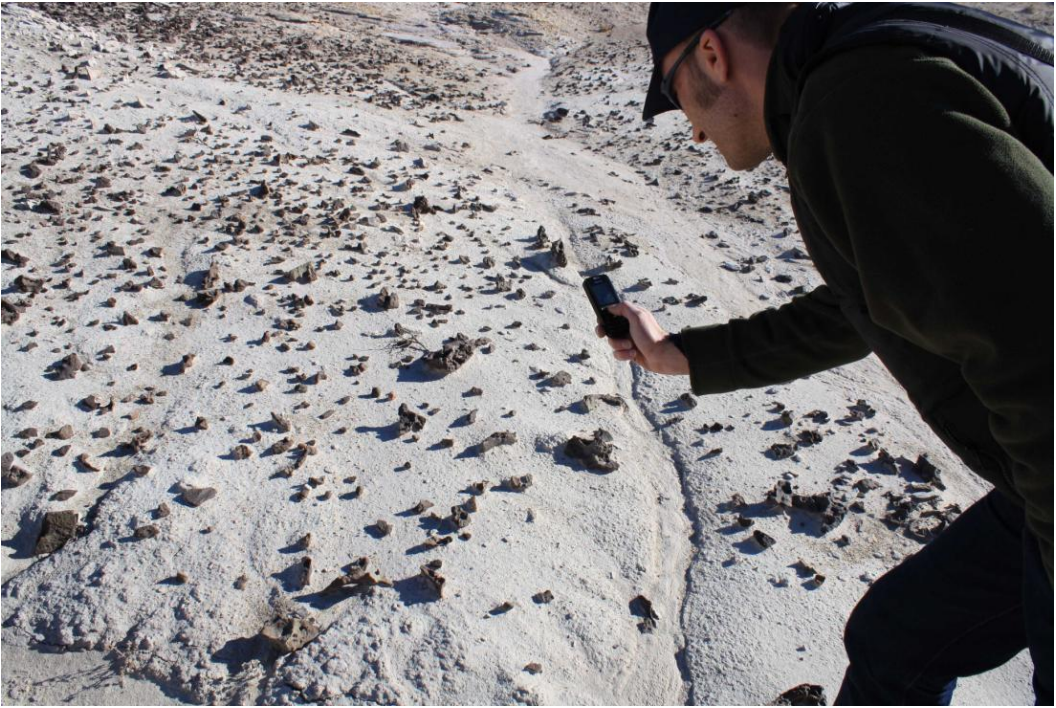

b)

**Figure 7:** Two further images of the Cyborg Astrobiologist using the Bluetooth-enabled Astrobiology Phone-cam in the Utah desert. Shading of the phone-cam screen was sometimes required, in order to view the details on the phone-cam screen.

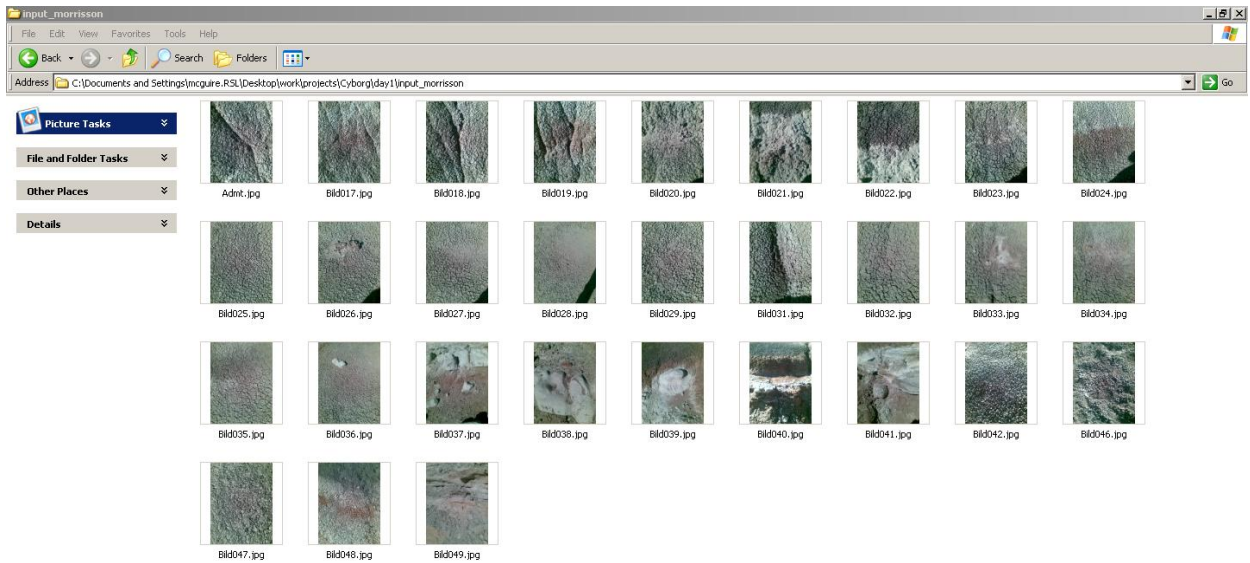

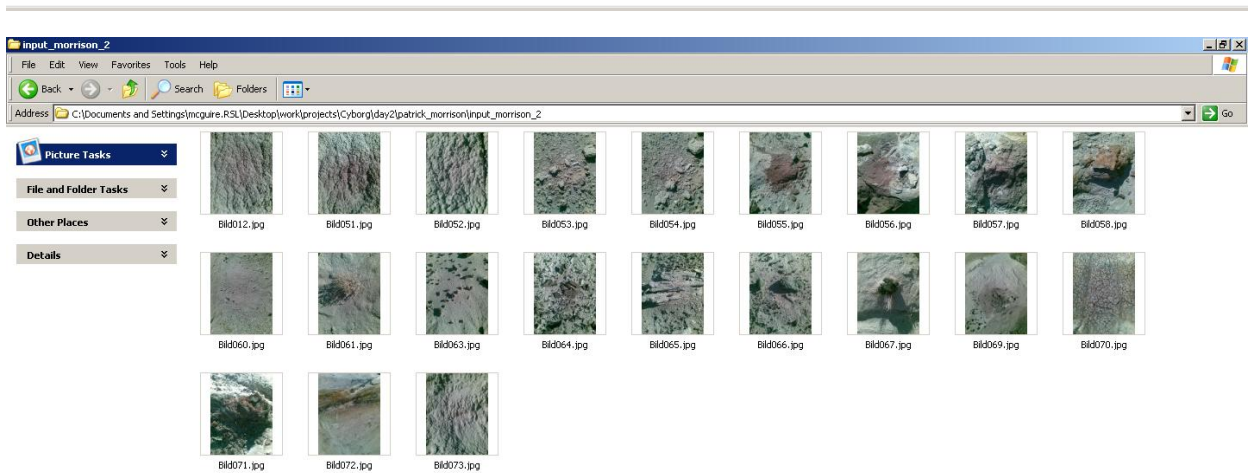

**Figure 8:** The sequence of images acquired during the first two days of the tests at the MDRS in Utah. Fig. 9 has the novelty maps for this image sequence, and Fig. 10 shows expanded views of several images in this sequence. See text for details.

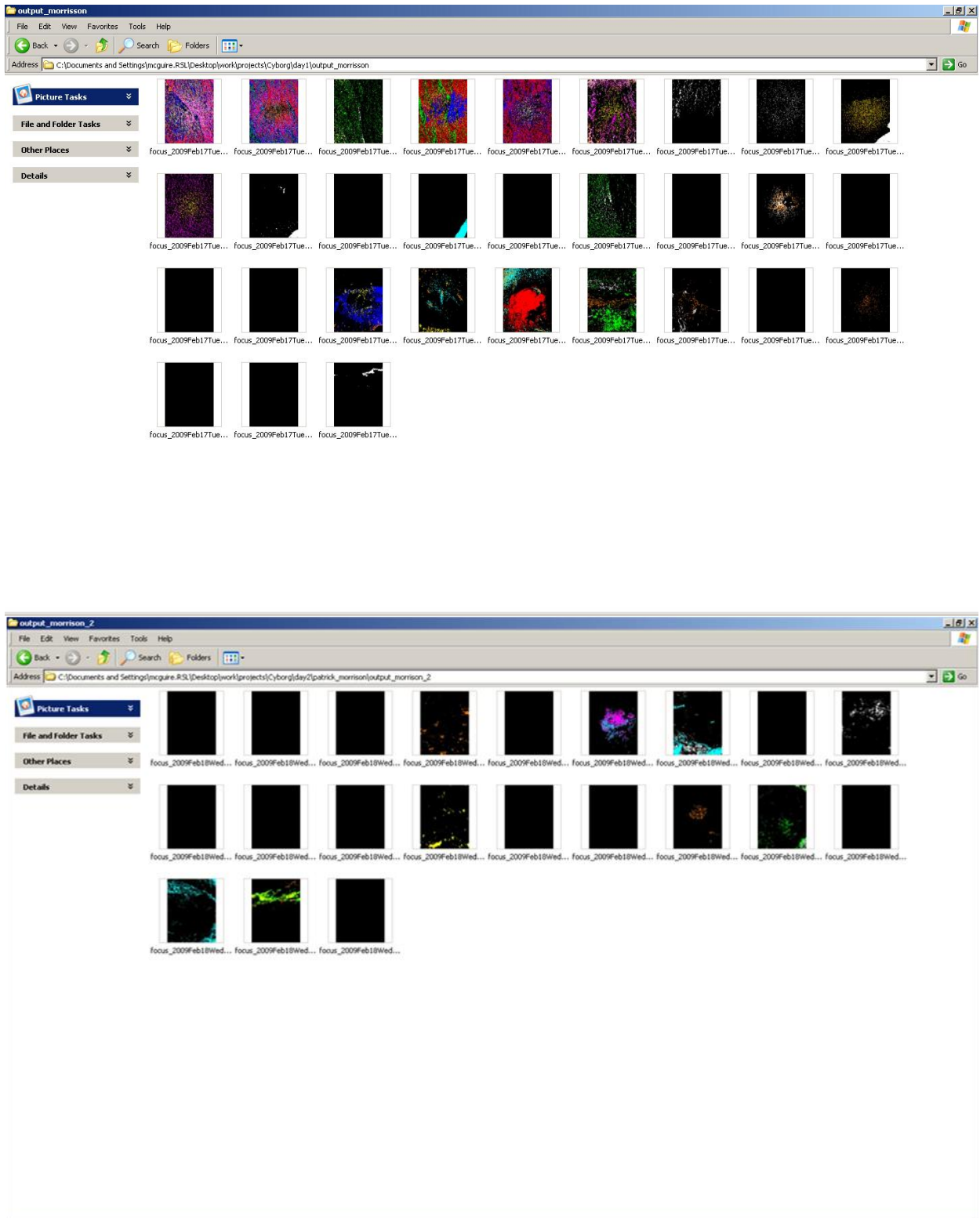

**Figure 9:** The sequence of novelty maps acquired during the first two days of the tests at the MDRS in Utah. These novelty maps correspond to the image sequence in Fig. 8. Fig. 10 shows expanded views of several images in this sequence. The colored regions are the novel regions in this image sequence. The black regions are the familiar regions in the image sequence. See text for details.

# Utah Tests Day#2: Novelty Detection

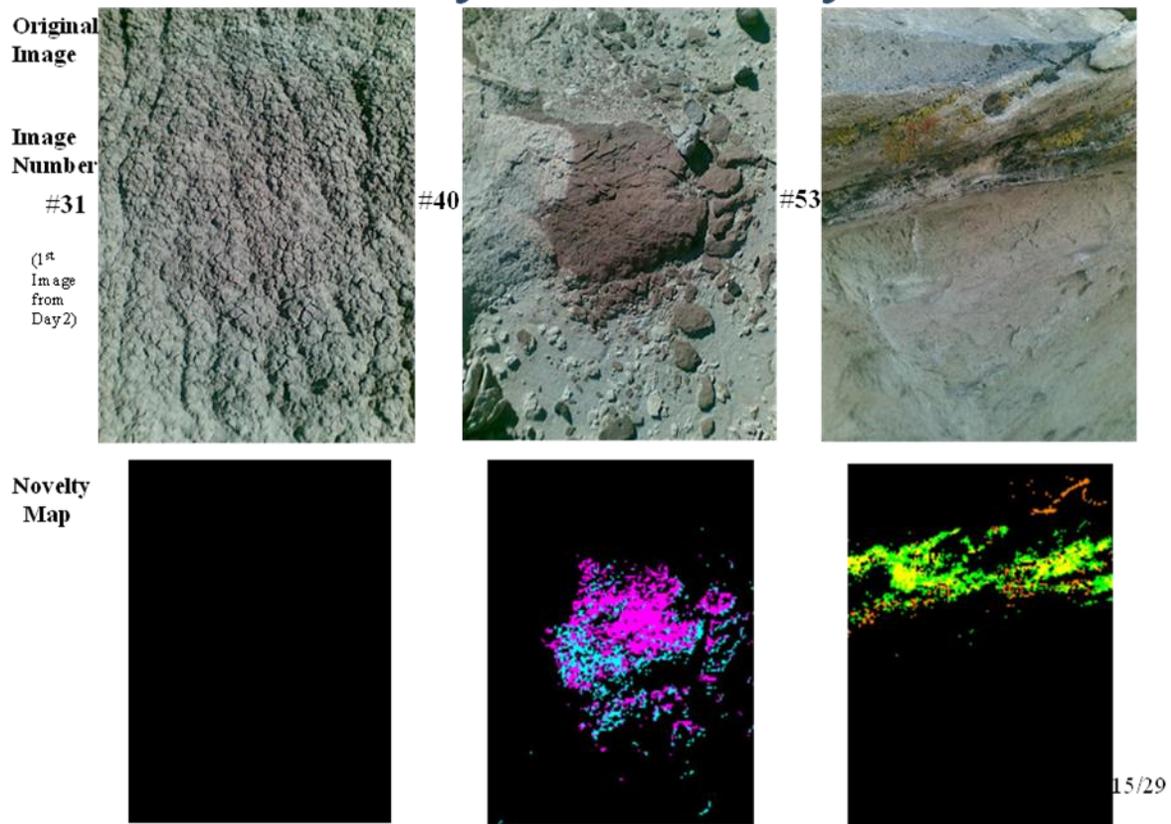

**Figure 10:** Examples of successful novelty detection during the field tests (Day #2). See The lower half of Figs. 8-9 for the sequencing. The system recognized the colors in image #31 as known and sends back a black image. Image #40 shows a mudstone outcrop, a color unknown to the system. The returned image indicates the novelty in cyan and magenta. Image #53 shows yellow to orange lichen, growing on sandstone. The system clearly indentifies the lichen as novel, whereas the surrounding rock is known and therefore colored black in the result image.

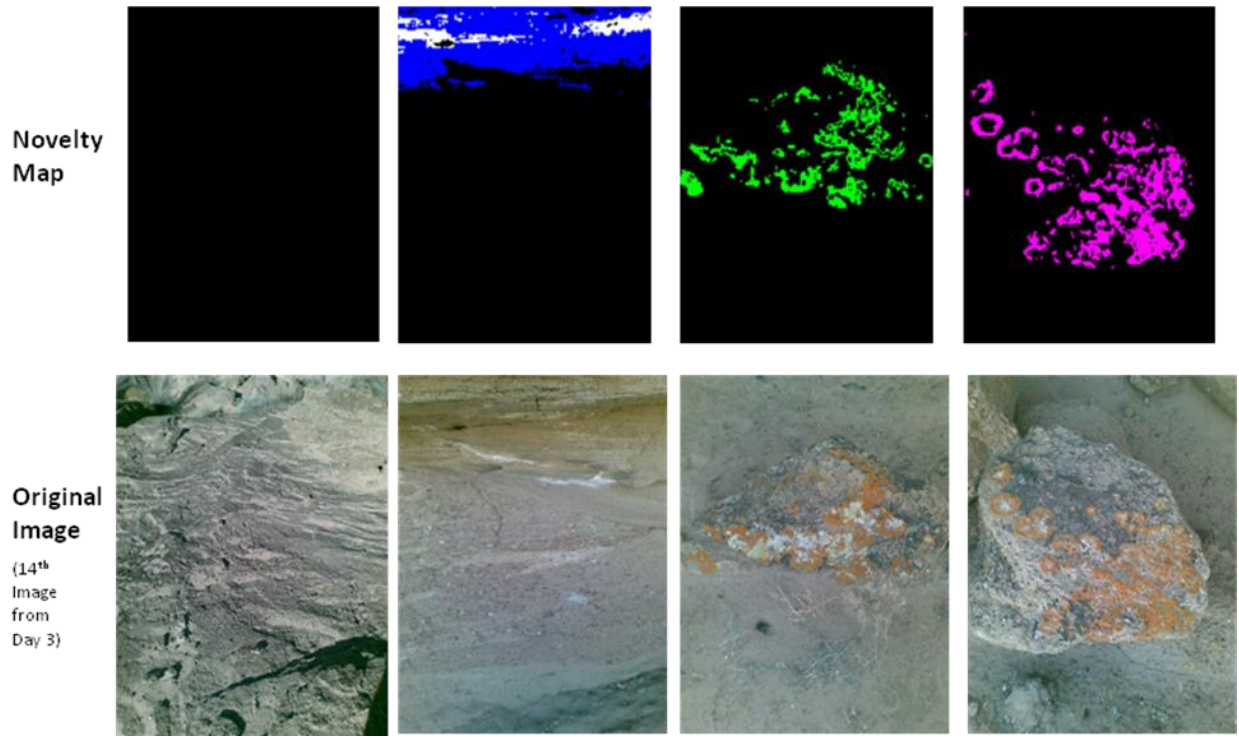

**Figure 11: (Bottom)** Zoom-in on four of the images from the image sequence, from Day #3 at the MDRS. The widths of these 4 images are about 2 m, 4 m, 0.4 m and 0.3m, from left to right. **(Top)** Zoom-in on the corresponding novelty maps (from Fig. 9). Note that the Cyborg Astrobiologist identified as familiar the 'normally' colored image on the left, and it identified as novel the yellow, orange and green colored lichen in the three images on the right. The system sometimes requires several images before it recognizes a color as being familiar.

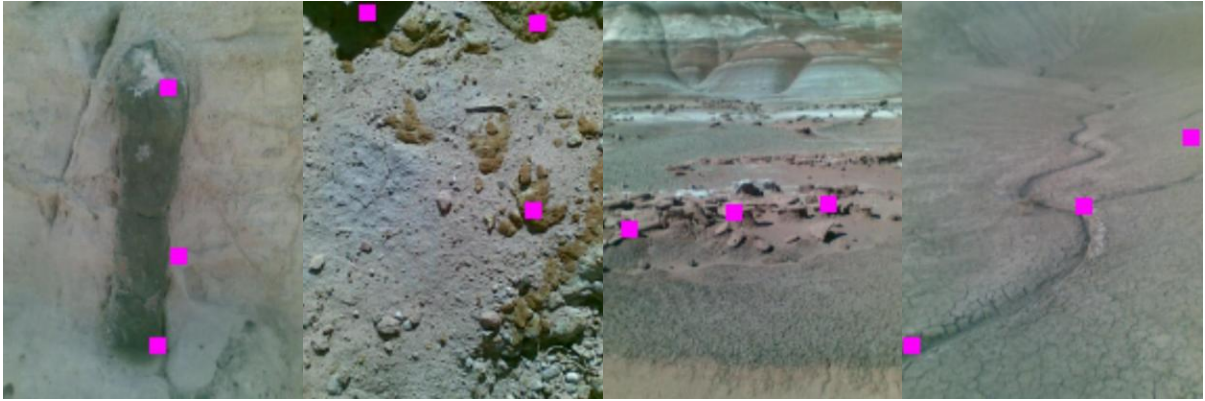

**Figure 12:** From left to right: Images 86, 87, 88 and 89 from day 3 of the MDRS test campaign. The widths of these images are approximately: 3 m, 0.3 m, 50 m, and 80 m, all measured half-way up the image. Three interest points (based upon uncommon mapping) are overlain upon each image in real-time and transmitted back to the phone-cam. Image 86 focuses on a piece of petrified wood. Image 87 focuses on a lichen. Image 88 focuses on a cluster of aeolian or water-eroded features. Image 89 focuses on a channel.